%% file: manuscript.tex
\documentclass[10pt,twocolumn,letterpaper]{article}

\usepackage{iccv}
\usepackage{times}
\usepackage{epsfig}
\usepackage{graphicx}
\usepackage{amsmath}
\usepackage{amssymb}
\usepackage{makecell}
\usepackage{multirow}
\usepackage{multicol}
\usepackage{booktabs}
\usepackage{pifont}

\usepackage{cite}
\usepackage{arydshln}

\usepackage[pagebackref=true,breaklinks=true,letterpaper=true,colorlinks,bookmarks=false]{hyperref}

\iccvfinalcopy % *** Uncomment this line for the final submission

\def\iccvPaperID{1536} % *** Enter the ICCV Paper ID here

\ificcvfinal\fi

\begin{document}

%%%%%%%%% TITLE
\title{Lip Reading for Low-resource Languages by Learning and Combining \\ General Speech Knowledge and Language-specific Knowledge}

\author{Minsu Kim\thanks{Both authors have contributed equally to this work.} \quad\quad Jeong Hun Yeo\footnotemark[1] \quad\quad Jeongsoo Choi \quad\quad Yong Man Ro\thanks{Corresponding author} \\
School of Electrical Engineering, KAIST, South Korea\\
{\tt\small \{ms.k, sedne246, jeongsoo.choi, ymro\}@kaist.ac.kr}\\
}

\maketitle
\ificcvfinal\thispagestyle{empty}\fi

%%%%%%%%% ABSTRACT
%%%%%%%%% ABSTRACT
\begin{abstract}
This paper proposes a novel lip reading framework, especially for low-resource languages, which has not been well addressed in the previous literature. Since low-resource languages do not have enough video-text paired data to train the model to have sufficient power to model lip movements and language, it is regarded as challenging to develop lip reading models for low-resource languages. In order to mitigate the challenge, we try to learn general speech knowledge, the ability to model lip movements, from a high-resource language through the prediction of speech units. It is known that different languages partially share common phonemes, thus general speech knowledge learned from one language can be extended to other languages. Then, we try to learn language-specific knowledge, the ability to model language, by proposing Language-specific Memory-augmented Decoder (LMDecoder). LMDecoder saves language-specific audio features into memory banks and can be trained on audio-text paired data which is more easily accessible than video-text paired data. Therefore, with LMDecoder, we can transform the input speech units into language-specific audio features and translate them into texts by utilizing the learned rich language knowledge. Finally, by combining general speech knowledge and language-specific knowledge, we can efficiently develop lip reading models even for low-resource languages. Through extensive experiments using five languages, English, Spanish, French, Italian, and Portuguese, the effectiveness of the proposed method is evaluated.
\end{abstract}
\vspace{-0.5cm}
%%%%%%%%% BODY TEXT
\section{Introduction}
It is a fascinating ability to understand the conversation by only looking at the speaker's lip movements without listening \cite{dodd1987lipreading}. If this were possible, we could easily hold conversations in crowded places, such as at concerts, and even with people who have trouble speaking up. With the great advance of deep learning, a technology called lip reading has made it possible to accurately infer what a speaker is saying without having to approach the speaker's voice. In recent years, the performance of lip reading has significantly improved from 60.1\% Word Error Rate (WER) to 26.9\% WER \cite{zhang2019spatio,shi2022learning} in LRS3 \cite{afouras2018lrs3}, a popular English benchmark database.

Such rapid progress could be made with large-scale audio-visual datasets \cite{chung2017lrw,yang2019lrw1000,chung2017lrs2,afouras2018lrs3,zhao2019CMLR}, improved neural network architecture \cite{assael2016lipnet,stafylakis2017combining,petridis2018end,afouras2018deep,xiao2020deformation,kim2021lip,ma2021end}, enhanced multi-modal learning strategies \cite{chung2017sync,zhao2020hearing,ma2021lira,kim2021cromm,kim2021multi,ren2021master,kim2022distinguishing,shi2022learning}, and carefully designed training methods \cite{feng2020learn,ma2021bornagain,ma2022multiple}. Among these progresses, self-supervised learning methods using audio-visual data show remarkable advancement in both audio-visual speech recognition and lip reading. Recently, AV-HuBERT \cite{shi2022learning} which pre-trains the transformer encoder with multi-modal inputs (\ie, audio and video) through masked prediction in a self-supervised manner, outperforms other previous lip reading methods once it is finetuned on lip reading tasks. Despite these advances, lip reading technologies have been developed primarily in English rather than in other languages. One main reason for this is the lack of enough labeled video-text paired data in other languages. For example, the popular lip reading dataset in English, LRS3 \cite{afouras2018lrs3}, consists of about 443 hours of video, while the available video-text paired dataset in Italian \cite{salesky2021mtedx} is only about 47 hours, which is not enough for the model to learn the characteristics of both lip movements and language. Therefore, to build lip reading models for other languages rather than English, a new method considering the insufficient training data should be developed.

In this paper, we focus on developing a novel lip reading method for low-resource languages which has not been well explored in previous literature. 
Specifically, we propose a novel training method for low-resource language lip reading that learns 1) general speech knowledge and 2) language-specific knowledge, and combines the two learned knowledge. First, general speech knowledge refers to the knowledge of modelling short-term speech that can be regarded as speech units (\ie, phonemes or visemes). Since different languages partially share common phonemes \cite{schultz2001multilingualASR,vu2014multilingual,luo2020multilingualVSR}, learning to model accurate speech units from high-resource language can be beneficial in modelling speech representations for low-resource language. To this end, we train the visual encoder to predict speech units from input lip movements through masked prediction using a high-resource language, English. Second, language-specific knowledge refers to the knowledge of translating learned speech representations into text, which can be regarded as the language modelling ability of a model. Since learning a language requires large-scale data \cite{devlin2018bert,yang2019xlnet,raffel2020T5}, it might be insufficient to only utilize the video-text paired data of low-resource language. To mitigate the problem, we propose a Language-specific Memory-augmented Decoder (LMDecoder) which can be trained from audio-text paired data in the target language and be applied for lip reading. The input of LMDecoder is set to speech units derived from audio, and Language-specific Memory (LM) saves language-specific audio features into memory banks, which are for transforming speech units into language-specific speech representations. Finally, after learning the two knowledge, we cascade the two modules (\ie, visual encoder and LMDecoder) and we can employ both the accurate lip movements modelling ability (\ie, general speech knowledge) of the visual encoder and the rich language modelling ability (\ie, language-specific knowledge) of LMDecoder, for low-resource language lip reading.

The effectiveness of the proposed method is evaluated with five languages, English (EN), Spanish (ES), French (FR), Italian (IT), and Portuguese (PT). Especially, English is utilized as a high-resource language so employed to learn general speech knowledge, and other languages are utilized as low-resource languages thus LMDecoder is trained on each low-resource language data. Through comprehensive experiments, we show the proposed method is effective in developing lip reading models not only for low-resource languages but also effective for the high-resource language, by achieving state-of-the-art performance on English data.
The contributions of this paper can be summarized as:
\begin{itemize}
\item To the best of our knowledge, this is the first attempt to analyze the effectiveness of different pre-training methods, self-supervised pre-training of encoders, supervised pre-training in a high-resource language, and pre-training of decoders with audio-text data, in building low-resource lip reading model.
\item We propose a novel method of learning and combining general speech knowledge and language-specific knowledge to effectively develop lip reading models for low-resource languages.
\item We conduct comprehensive experiments with five languages, English, Spanish, French, Italian, and Portuguese, and we show the effectiveness of the proposed method in developing lip reading models for different nationalities, even with a small-scale dataset.
\end{itemize}  

\section{Related Work}
\subsection{Lip reading}
Lip reading \cite{zhao2020mutual,zhang2020facecutout,hong2021speech,kim2023lip, choi2023intelligible,ma2023auto} aims to predict the speech content by watching talking face videos only. Along with the advancement of Deep Learning and speech processing technology, lip reading technology achieves significant development. Early work \cite{chung2017lrw} proposed a lip reading model consisting of CNN to predict word from word-level English data. \cite{stafylakis2017combining, petridis2018end} proposed an architecture composed of ResNet \cite{he2016deep} and RNN \cite{hochreiter1997lstm,chung2014gru} to improve the word-level lip reading performances.
\cite{weng2019twostream,xiao2020deformation} proposed to use optical flow information with RGB information by encoding them with two-stream networks. \cite{martinez2020mstcn} changed the RNN-based back-end architecture with temporal convolutions and achieved significant performance improvement in word-level lip reading. Besides the word-level lip reading, \cite{assael2016lipnet} proposed an end-to-end sentence-level lip reading model that utilizes Connectionist Temporal Classification (CTC) \cite{graves2012ctc}. Sentence-level large-scale audio-visual datasets, LRS2 \cite{chung2017lrs2} and LRS3 \cite{afouras2018lrs3}, are proposed to boost lip reading research. By adopting transformer \cite{vaswani2017attention}, powerful architecture for modelling sequence data, \cite{afouras2018deep} significantly improved the sentence-level lip reading performances. Recently, transformer-variants architectures \cite{ma2021end,hong2022visual,prajwal2022sub,hong2023watch} are shown promising lip reading and audio-visual speech recognition performances. There are other works that tried to enhance lip reading performances by focusing on developing training strategies. \cite{ma2021bornagain,ren2021master,afouras2020asr,zhao2020hearing} employed knowledge distillation \cite{hinton2015distilling} to bring knowledge of the superior model into the student model. \cite{kim2021multi,kim2021cromm,kim2022distinguishing,yeo2023multi,yeo2023akvsr} proposed to use memory networks to use the auditory knowledge in lip reading without audio inputs. \cite{almajai2016improved,kim2022speaker,kim2023prompt} handled the speaker-dependency issue and proposed speaker-adaptive lip reading models. Recently, pre-training neural networks using self-supervised training methods showed remarkable lip reading performances \cite{chung2017sync,ma2021lira,shi2022learning}.

%%%%%%%%%%%%%%%%%% Fig.1 %%%%%%%%%%%%%%%%%%
\input{Figure/Figure_1.tex}
%%%%%%%%%%%%%%%%%%%%%%%%%%%%%%%%%%%%%%%%%%%

However, most of the previous research is focused on developing lip reading models in principal languages, such as English and Mandarin. Lip reading for different languages, especially low-resource languages, has not been well explored \cite{ma2022multiple}. In this paper, we propose a new method for low-resource languages that contain a small-scale visual-text paired dataset. By learning and combining general speech knowledge and language-specific knowledge, the proposed method can effectively learn how to model the lips and the target language, even for the low-resource language.

\subsection{Pre-training strategies}
\vspace{-0.2cm}
Recently, in diverse areas, the pre-trained model shows remarkable performances when they are applied to different downstream tasks \cite{devlin2018bert,li2019vlp1,chen2020vlp2,zhou2020vlp3,zhang2020vlp4,dou2022vlp5,sun2019vlp6,radford2021vlp7}. It is also shown remarkable performances in the speech recognition area. In audio speech modelling, wav2vec2.0 \cite{baevski2020wav2vec} and HuBERT \cite{hsu2021hubert}, proposed to learn the speech representations by predicting speech units obtained by clustering the acoustic features (\eg, MFCC). They achieved state-of-the-art speech recognition performances by pre-training the model on large-scale unlabeled data. In visual speech modelling, \cite{chung2017out,ma2021lira,shi2022learning} proposed self-supervised pretraining methods using audio-visual correspondences or masked prediction similar to audio pre-training methods. By finetuning the pre-trained model to the lip reading task, they achieved better lip reading performances than the trained model from the scratch.

In this paper, we also try to pre-train the visual encoder to learn general speech knowledge by predicting speech units from lip video using high-resource languages. Moreover, to learn language-specific knowledge which will be utilized to translate the captured speech units into words, we propose Language-specific Memory-augmented Decoder (LMDecoder) which can be pre-trained on audio-text paired data. 

\subsection{Vector quantization}
Since discrete representations are natural to express many modalities, using discrete representations in Deep Learning shows great progresses in diverse areas, such as image generation \cite{van2017neuraldiscrete,esser2021vqgan,gu2022vector} and speech processing \cite{baevski2020wav2vec,lakhotia2021gslm,hsu2021hubert,baevski2019vq,polyak2021speech,chang2023exploration,kim2023many}. Especially, by discretizing audio using vector quantization, we can obtain discriminative hidden units which are highly correlated with the acoustic units (\ie, phoneme) \cite{hsu2021hubert}. We try to use the speech units obtained through vector quantization of input video and audio in learning general speech knowledge and language-specific knowledge.

\section{Method}
Our objective in this paper is to develop lip reading models for low-resource languages. Different from English, other languages (\eg, Italian, French, Korean, Japanese, etc.) have smaller video-text paired data for developing lip reading networks. Therefore, lip reading research has been mainly focused on English. To mitigate the insufficient visual-text paired data of the low-resource language in building a lip reading model, we propose to learn 1) general speech knowledge from a high-resource language and 2) language-specific knowledge from audio-text paired data. 

\subsection{Learning general speech knowledge}
\label{sec:3.1}
It is known that different languages share some common phonemes \cite{schultz2001multilingualASR,vu2014multilingual,luo2020multilingualVSR,kim2023many}, which means that knowledge learned to model speech units from lip movements in one language can be extended to other languages. Therefore, to effectively learn to model the lip movements of low-resource language, we try to bring the knowledge of a pre-trained model that is trained to model the speech units from a high-resource language, English. Motivated by the recent success of learning speech representations by predicting speech units in a self-supervised manner \cite{baevski2020wav2vec,hsu2021hubert,shi2022learning,shi2022robust}, we train the visual encoder with masked prediction to learn general speech knowledge.

Specifically, the contiguous $\alpha$ frames of input video $x_v$ with $T$ frames are masked out. Then the masked video $\Tilde{x}_v$ is encoded through a visual front-end and a transformer to produce visual features $f_v$. Then, the visual encoder (\ie, visual front-end and transformer) is guided to predict the speech units of the masked region indicated by an indicator $M_t\in\{0,1\}$, where the value $1$ indicates $t$-th frame is masked while the value $0$ indicates not. The target speech units $z_t\in\{1,\dots,C\}$ with $C$ classes are obtained by quantizing Mel-frequency Cepstral Coefficient (MFCC) of the audio corresponding to the input video using a discrete latent variable model (\eg, K-means), which will be iteratively improved by using the learned features instead of MFCC similar to \cite{baevski2020wav2vec,hsu2021hubert,shi2022learning}. The process of learning general speech knowledge can be written as follows,
\begin{align}
    \mathcal{L}_{GSK} = - \sum_{\{t|M_t=1\}}z_t\log(\hat{z}_t),
\end{align}
where $\hat{z}_t=\text{Softmax}(F(f_v^t))$ is the probability of the predicted speech unit using a classifier $F(\cdot)$. For the implementation, we follow the recent observation of \cite{shi2022learning} that using both audio and video inputs to learn the speech representations is better than utilizing the video inputs only, and we use the multi-modal inputs. By training the visual encoder with the masked prediction of speech units on the large-scale dataset, the visual encoder can embed lip video into discriminative speech representations, which will be extended to other languages. The process for learning general speech knowledge is illustrated in Fig.\ref{fig:1}a.

\subsection{Learning language-specific knowledge}
The final goal of lip reading is translating the captured lip movements into words, which implies that the ability of language modeling can largely affect the final performance. However, for the low-resource language, video-text paired data might be insufficient for the model to learn to construct language. To handle this, as shown in Fig.\ref{fig:1}b, we propose a Language-specific Memory-augmented Decoder (LMDecoder) which can learn language-specific knowledge from audio-text paired data usually richer than video-text data.

Specifically, LMDecoder includes Language-specific Memory (LM) that can save speech representations of the target language according to speech units. Therefore, after training, we can extract language-specific speech representations from LM by examining the input speech units. When input audio is given, it is quantized to speech units $x_a$ having $C$ classes, similar to the obtaining of $z_t$ in Sec.\ref{sec:3.1}. By quantizing the input audio into speech units, the learned general speech knowledge can be naturally fit to be utilized in LMDecoder when the visual encoder and LMDecoder are combined. Then, LM converts the input speech units into language-specific audio features $f_a$ by accessing memory banks $B\in\mathbb{R}^{C\times d}$ corresponding to speech units as follows,
\begin{align}
    f_a^t = B_i \quad \text{such that} \quad x_a^t=i,
\label{eq:2}
\end{align}
where $d$ is the dimension of audio features. This procedure is illustrated in Fig.\ref{fig:2} and it is similar to accessing the codebook in \cite{esser2021vqgan} and also to using auditory features in lip reading using the memory network of \cite{kim2021cromm,kim2021multi,kim2022distinguishing,yeo2023akvsr}. Then, a decoder translates the audio features into words in an autoregressive manner \cite{sutskever2014sequence}. Let $y$ be the ground-truth text tokens, then the process of learning language-specific knowledge of LMDecoder can be written as follows,
\begin{align}
    \mathcal{L}_{LSK}=-\log p(y|x_a)
\end{align}
where $p(y|x_a)=\Pi_{j=1}^Jp(y_j|y_{< j},x_a)$ and $J$ represents the length of text tokens. As the learning of language-specific knowledge is purely available with audio-text paired data, the LMDecoder can learn to model the target language from large-scale data, even if the video-text paired dataset is small for the target language. Moreover, since the saved language-specific representations in LM are auditory features, we can bring the rich speech information of audio into lip reading, similar to \cite{kim2021cromm,kim2021multi,kim2022distinguishing,yeo2023akvsr}.

%%%%%%%%%%%%%%%%%% Fig.2 %%%%%%%%%%%%%%%%%%
\input{Figure/Figure_2.tex}
%%%%%%%%%%%%%%%%%%%%%%%%%%%%%%%%%%%%%%%%%%%

\subsection{Lip reading for low-resource language}
After training the visual encoder to have general speech knowledge and the LMDecoder to have language-specific knowledge, we combine the two modules to compose the lip reading pipeline for low-resource language (Fig.\ref{fig:1}c). To access the saved language-specific audio features in LM, we employ scaled dot-product attention of \cite{vaswani2017attention} using the encoded visual features $f_v$. Through attention, the potential mismatches between speech units predicted from video and predicted from audio can be minimized when accessing the memory banks. Specifically, when visual features $f_v$ are encoded by the visual encoder, language-specific audio features saved in LM (\ie, $B$) are retrieved through an attention mechanism as follows, 
\begin{equation}
\begin{gathered}
    Q^t = f_v^tW_q, \quad K = BW_k, \quad V = BW_v, \\
    f_a^t = \text{Softmax}(\frac{Q^tK^\top}{\sqrt{d}})V,
\end{gathered}
\label{eq:4}
\end{equation}
where $W_q$, $W_k$, and $W_v$ are embedding matrices for query, key, and value, respectively. By using visual features as a query, we can access the memory banks $B$ of LM to find and extract language-specific audio features related to the input lip movements. This also can be viewed as a soft attention \cite{xu2015show} version of Eq. (\ref{eq:2}). Finally, with the extracted language-specific audio features, LMDecoder predicts text tokens of the target language, $\hat{y}$, in an auto-regressive manner by utilizing the learned language-specific knowledge.

\section{Experimental Setup}
\subsection{Network architecture}
Basically, the visual encoder has the same architecture as that of the AV-HuBERT Base \cite{shi2022learning} except for the LRS2 experiment which utilizes AV-HuBERT Large configuration. It is composed of a visual front-end and transformer encoders. Specifically, the visual front-end is comprised of ResNet18 \cite{he2016deep} whose first stem layer is modified with 3D convolution \cite{petridis2018end}. The transformer has 12 encoder layers where each encoder has a 768 embedding dimension (\ie, $d=768$), a 3,072 feed-forward dimension, and 12 attention heads. The LMDecoder consists of Language-specific Memory (LM), transformer encoders, and transformer decoders. LM has memory banks with an embedding matrix size of $C\times d$, where $C$ is set to 1,000. The transformer encoder has 4 layers to model the context from the extracted audio features $f_a$, with the same embedding size as the transformer in the visual encoder. The transformer decoder has 6 layers with a 768 embedding dimension, a 3,072 feed-forward dimension, and 4 attention heads, to predict the text tokens. To bridge the visual encoder and LMDecoder, we utilize scaled dot-product attention of \cite{vaswani2017attention} and the size of each embedding layer is set to the dimension of audio features (\ie, $W_q,W_k,W_v\in\mathbb{R}^{d\times d}$).

For obtaining the target speech units $z$, we use that of \cite{shi2022learning}; they are obtained by clustering MFCC at the first iteration, and then changed to the cluster of learned audio-visual representations through the iteration. For obtaining the speech units $x_a$ from the input audio, we use the features encoded at the 11-th layer of a pre-trained HuBERT \cite{hsu2021hubert} that trained on VoxPopuli \cite{wang2021voxpopuli} and perform K-means clustering.

\subsection{Implementation details}
Our experiments are implemented using an open-source toolkit, fairseq \cite{ott2019fairseq}. For the video input, the lip region is detected through face detection \cite{li2018dsfd} and landmark detection \cite{lee2019lightweight} and we crop the region with a size of 96$\times$96. Every input frame is converted into grayscale. For data augmentation purposes, horizontal flipping and random cropping into a size of $88\times88$ are applied to the visual inputs during training. For masked prediction of speech units, $\alpha$ is set to 5 at the last iteration and masking is performed by substituting with random contiguous frames of the same video, following \cite{shi2022learning}. For training LMDecoder, the audio input is resampled to 16kHz and quantized through the aforementioned pre-trained HuBERT \cite{hsu2021hubert}. For the text tokenizer, we use a subword-level tokenizer, sentencepiece \cite{kudo2018sentencepiece}, and set the dictionary size to 1,000 for all languages.

The visual encoder and LMDecoder are pre-trained separately. The visual encoder is trained on audio-visual data of a high-resource language, English, and we directly utilize the pre-trained model of \cite{shi2022learning} for the visual encoder. LMDecoder is trained on audio-text paired data in the target language. For each language (\ie, ES, FR, IT, and PT), we use the data corresponding to the target language from Multilingual LibriSpeech (MLS) \cite{Pratap2020MLS}, the audio-text paired dataset, to train LMDecoder. After training the visual encoder and LMDecoder, we combine them with an attention layer and finetune the entire model on the lip reading data (\ie, video-text paired data) of the target language. For the objective function to train LMDecoder and finetuning the entire lip reading models on the target language video-text paired data, we use hybrid CTC/attention loss \cite{watanabe2017hybrid}. For decoding, we do not use an external language model and the joint CTC/attention decoding, and only utilize the output of the decoder for all experiments.

For pre-training LMDecoder and finetuning the entire lip reading model in the target languages, we employ Adam \cite{kingma2014adam} optimizer and tri-stage learning rate schedules for all experiments. The peak learning rate is set to 0.001, and the warmup stages for pre-training and finetuning are set to 15,000 steps and 10,000 steps, respectively. LMdecoder is trained for 60,000 steps on audio-text paired data of the target language. For finetuning the lip reading model, we train the model for 50,000 steps by using video-text paired data in the target language, except for English. We train the English lip reading model for 30,000 steps by freezing the visual encoder for 20,000 steps. Further details can be found in the supplementary material.                

\subsection{Dataset}
{\bf Multilingual TEDx (mTEDx)} \cite{salesky2021mtedx} is a multilingual TEDx corpus for speech recognition and translation. The dataset is composed of speech audio and transcriptions, for 8 languages. In order to use the dataset in lip reading, we download the video from Youtube by using the links provided by the dataset. Based on the data splits of the dataset, we follow \cite{ma2022multiple} to remove the video not containing a speaker and unavailable video online. We utilize Spanish (ES), French (FR), Italian (IT), and Portuguese (PT) to evaluate the proposed method. The dataset size of each language is represented in Table \ref{table:1}.

{\bf Multilingual LibriSpeech (MLS)} \cite{Pratap2020MLS} dataset is a large multilingual audio-text paired dataset for Audio-based Speech Recognition (ASR). The dataset is derived from audiobooks and consists of 8 languages. We utilize ES, FR, IT, and PT languages to train the proposed LMDecoder to learn language-specific knowledge. The dataset size of each language is represented in Table \ref{table:1}. Please note that the available audio-text paired data is much larger than video-text paired data (\ie, mTEDx).

%########### Table 1 #############
\input{Table/Table1.tex}
%#################################

{\bf LRS3} \cite{afouras2018lrs3} is a large-scale English sentence-level audio-visual dataset. It consists of about 439 hours of videos. We use 433 hours of training data to pre-train the visual encoder with masked predictions of speech units, for learning general speech knowledge.

{\bf VoxCeleb2} \cite{chung2018voxceleb2} is a large-scale unlabeled audio-visual dataset. It consists of about 2,442 hours of videos. We use 1,326 hours of training data following \cite{shi2022learning} to pre-train the visual encoder along with the LRS3 dataset.

{\bf LRS2} \cite{chung2017lrs2} is another large-scale English sentence-level audio-visual dataset derived from television shows. It has about 224 hours of data. We use the dataset to evaluate the effectiveness of the proposed lip reading framework in the high-resource language, English.

\subsection{Baselines}
\label{sec:4.4}
In order to analyze the effectiveness of the proposed method in developing lip reading models for low-resource languages, we set five baselines to be compared. All the methods are implemented with the same settings.

{\bf Supervised pre-training.} This baseline is to evaluate whether a well-trained lip reading model in a high-resource language, English, can be employed for other languages. To this end, we pre-train a state-of-the-art lip reading model, \textbf{CM-seq2seq} \cite{ma2021end}, on large-scale labeled datasets in English, a total amount of 814 hours composed of LRW \cite{chung2017lrw}, LRS2 \cite{chung2017lrs2}, and LRS3 \cite{afouras2018lrs3}. Then, the entire pre-trained model is directly finetuned on each target language.

{\bf Self-supervised pre-training of encoders.} This baseline is to evaluate whether the learned general speech knowledge, the ability to model speech units from lip movements, is beneficial when it is applied to other languages. To this end, we only utilize the visual encoder pre-trained in a high-resource language (\ie, 1,759 hours of English data) with the objective of masked prediction of speech units. Then, a decoder, to be trained from scratch, is attached to the visual encoder to construct the lip reading pipeline and trained on the target lip reading dataset. The model is trained on a total of 50K iterations and the pre-trained visual encoder is frozen until 20K iterations. Since this method can be viewed as the application of AV-HuBERT \cite{shi2022learning} in different languages, we denote this method as \textbf{AV-HuBERT}.

{\bf Pre-training of decoders.} This baseline is to evaluate the effectiveness of pre-training of decoders on audio-text paired data. To this end, we pre-train a decoder through ASR task on each target language data of MLS \cite{Pratap2020MLS} by attaching it to a pre-trained audio encoder of \cite{shi2022learning}. After training, the decoder is attached to the pre-trained visual encoder and the entire model is finetuned on the target lip reading dataset. We denote this method as \textbf{ASR Pre-train}. This method can be viewed as the absence of LM and quantized speech units in the proposed method.

{\bf Distillation of pre-trained knowledge.} This baseline is to evaluate the effectiveness of the knowledge distillation-based method of \cite{ma2022multiple} in low-resource language lip readings. To this end, we first pre-train both lip reading and ASR models initialized from AV-HuBERT \cite{shi2022learning} on the target lip reading dataset. Then, by utilizing the pre-trained lip reading and ASR models as teachers, a new lip reading model initialized from AV-HuBERT is trained by distilling the knowledge of the two teachers. We follow other training configurations of \cite{ma2022multiple} to train the model, and denote this method as \textbf{Auxiliary Task}.

{\bf Proposed Method.} The final method is the proposed method that utilizes the pre-trained general speech knowledge and language-specific knowledge. To evaluate the effectiveness of the proposed method, we pre-train the visual encoder in a high-resource language and LMDecoder on the target language data from MLS \cite{Pratap2020MLS}. Then, the two modules are attached by using an attention layer and the entire model is finetuned on the target lip reading dataset.

%########### Table 2 #############
\input{Table/Table2.tex}
%#################################

\section{Experimental Results}
\subsection{Comparison with the state-of-the-art methods}
Before evaluating the lip reading performances for the low-resource languages, we first evaluate the effectiveness of the proposed framework on a high-resource language dataset, LRS2 \cite{chung2017lrs2}. To this end, we train our LMDecoder with the combination of training datasets of LRS2 and LRS3. Then, the LMDecoder is attached to the pre-trained visual encoder and finetuned on LRS2 dataset. The evaluation results on LRS2, are shown in Table \ref{table:2}. We compare the performances obtained by using `video-text data' of LRS2 only, if some works utilize extra video-text data, we report the performance obtained by using minimum extra video-text data. The proposed method outperforms the previous state-of-the-art methods and sets a new state-of-the-art performance, by achieving 23.8\% WER. In particular, the proposed method outperforms AV-HuBERT \cite{shi2022learning} that shares the same visual encoder by 1.7\% WER, which means that the proposed LMDecoder can contribute to even high-resource language lip-reading by enriching language modeling ability.

\subsection{Effectiveness in low-resource languages}
To evaluate the effectiveness of the different methods on low-resource lip reading, we compare the performances of five different lip reading methods described in Sec. \ref{sec:4.4} on four low-resource languages, ES, FR, IT, and PT. Table \ref{table:3} shows the comparison results on mTEDx-IT, Table \ref{table:4} shows results on mTEDx-FR, Table \ref{table:5} shows results on mTEDx-ES, and Table \ref{table:6} shows results on mTEDx-PT.

\textbf{Effectiveness of learning general speech knowledge.} Firstly, we compare \textit{CM-seq2seq} and \textit{AV-HuBERT} to confirm whether learning lip reading in a high-resource language using large-scale labeled video-text paired data is better or learning general speech knowledge from a high-resource language is better for low-resource languages lip reading. \textit{CM-seq2seq} is pre-trained on the lip reading task using 814 hours of English video-text data and then finetuned on each target language, while \textit{AV-HuBERT} is pre-trained on the speech units prediction task using 1,759 hours of English audio-visual data and then finetuned on each target language. \textit{CM-seq2seq} achieves 88.41\% WER and \textit{AV-HuBERT} achieves 77.36\% WER, on French shown in Table \ref{table:4}. The results indicate that even if \textit{CM-seq2seq} is utilized large-scale English labeled data, the knowledge cannot be fully transferred for French lip reading. On the other hand, \textit{AV-HuBERT}, which only utilizes labeled data of French, achieves better results by expanding the general speech knowledge learned from English data into French. Similar tendencies can be found in other languages, IT, ES, and PT in Tables \ref{table:3}, \ref{table:5}, and \ref{table:6}. The results indicate that it would be beneficial to learn how to encode general speech units instead of learning to translate the lips into text in a high-resource language, for the purpose of adapting the pre-trained model to low-resource languages.

%########### Table 3 #############
\input{Table/Table3.tex}
%#################################

%########### Table 4 #############
\input{Table/Table4.tex}
%#################################

\textbf{Effectiveness of pre-training the decoder.} In order to evaluate the effectiveness of pre-training the decoder using audio-text data, we compare \textit{AV-HuBERT} and \textit{ASR Pre-train}. The decoder of \textit{AV-HuBERT} is trained from scratch by using the video-text paired data of the target language, while the decoder of \textit{ASR Pre-train} is trained after initializing with the pre-trained model on audio-text paired data of the target language. Since the visual encoders of the two models are the same, we can focus on the effects of pre-training the decoder using audio-text paired data. As shown in Table \ref{table:5}, the performance of \textit{AV-HuBERT} on Spanish is 71.68\% WER while \textit{ASR Pre-train} achieves 70.80\% WER. The results confirm that even if we learn general speech knowledge from a large-scale English dataset, the ability to model language might be insufficient to be learned from the small-scale target language dataset (\ie, video-text paired dataset). By adapting the language knowledge learned from an audio-text paired dataset which is larger than the video-text paired dataset, we can improve the lip reading performances for the low-resource languages. Results for other languages, IT, FR, and PT, are shown in Tables \ref{table:3}, \ref{table:4}, and \ref{table:6}.

\textbf{Effectiveness of learning language-specific knowledge through LMDecoder.} In order to evaluate the effectiveness of the proposed method of learning language-specific knowledge through LMDecoder, we compare \textit{ASR Pre-train} and \textit{Proposed Method}. Different from \textit{ASR Pre-train}, the \textit{Proposed Method} is additionally trained to save language-specific audio features in LM and to construct the mapping between speech units and language-specific audio features by using quantized speech units. Therefore, by comparing the two methods, we can validate the effectiveness of the proposed lip reading framework for low-resource languages. \textit{ASR Pre-train} achieves 71.28\% WER on Italian while the \textit{Proposed Method} achieves 68.04\% WER which outperforms \textit{ASR Pre-train} by about 3.24\% WER, shown in Table \ref{table:3}. Since the proposed LMDecoder can transform the encoded visual features (\ie, speech units) into language-specific audio features, it can fully utilize the learned general speech knowledge of the pre-trained visual encoder, when the two pre-trained modules (\ie, visual encoder and LMDecoder) are combined. Moreover, as the LM can provide rich speech information of audio by reading the memory banks, we can also employ the complementary effects of multi-modality as proven to be effective for lip reading in previous works \cite{kim2021multi,kim2021cromm,kim2022distinguishing}. Similar tendencies can be found in tables \ref{table:4}, \ref{table:5}, and \ref{table:6}.
 
\textbf{Comparison with distillation-based method.}
Finally, we compare the lip reading performance with a distillation-based method that utilizes knowledge distillation as an auxiliary task. In Tables \ref{table:3}, \ref{table:4}, \ref{table:5}, and \ref{table:6}, compared to \textit{Auxiliary Task}, the \textit{Proposed Method} outperforms the method in all languages. Even if \textit{Auxiliary Task} tried to learn from using the knowledge of superior models (\ie, pre-trained lip reading and ASR models), the \textit{Proposed Method} can achieve better performance by employing language-specific knowledge learned from a larger audio-text paired dataset. 

Comparing the performances of \textit{Proposed Method} with other baselines, we can confirm the effectiveness of the proposed lip reading framework for low-resource languages.

%########### Table 5 #############
\input{Table/Table5.tex}
%#################################

%########### Table 6 #############
\input{Table/Table6.tex}
%#################################

%########### Table 7 #############
\input{Table/Table7.tex}
%#################################

%########### Table 8 #############
\input{Table/Table8.tex}
%#################################

\subsection{Ablation study}
\textbf{Different audio-text paired datasets.} We perform ablation studies to examine the effectiveness of the proposed lip reading framework. Firstly, we examine the effect of different audio-text paired datasets in learning language-specific knowledge of LMDecoder. We pre-trained three variants of LMDecoder by using MLS, mTEDx, and both datasets. Then, each model is attached to the pre-trained visual encoder, and the entire model is finetuned on the target language lip reading dataset. For the ablation study, we use Italian datasets (\ie, MLS-IT and mTEDx-IT). The ablation results are shown in Table \ref{table:7}. MLS-IT dataset has 247 hours of training data and mTEDx-IT dataset has 47 hours of training data. Using MLS-IT only to train LMDecoder achieves 70.45\% WER. By using an extra audio-text paired dataset, MLS-IT, to train language-specific knowledge for LMDecoder, we can improve the performance from the baseline that uses the scratch decoder (\ie, 73.24\% WER) by 2.79\% WER. Moreover, by using audio-text paired data of mTEDx-IT only, we can still improve the performance and achieve 71.47\% WER. This shows the effectiveness of the LM in providing the saved language-specific audio features corresponding to speech units. By using both datasets, the performance improves to 68.04\% WER, which shows the effectiveness of learning language-specific knowledge using audio-text paired data in building low-resource language lip reading models.

\textbf{Effectiveness of Language-specific Memory (LM).} To evaluate the effectiveness of Langauge-specific Memory (LM) in LMDecoder, we experiment by eliminating the LM from the proposed method. Therefore, the decoder of \textit{Without LM} model is trained with quantized speech units but the LM is not included. The performance of \textit{Without LM} on Italian is shown in Table \ref{table:8}. By eliminating the proposed LM, the lip reading performance is degraded by about 3\% WER. The result clearly indicates that the saved language-specific audio features in LM can provide beneficial information when it is combined with general speech knowledge, with the following two roles; 1) constructing mapping between speech units and language-specific audio features, and 2) providing rich auditory information which can complement the lip reading model.

\textbf{Different amounts of video-text data.} In order to investigate the effectiveness of the proposed method under different amounts of video-text data situations, we experiment with 1/3 (15.7h), 2/3 (31.3h), and all (47h) of the video-text data of mTEDx-IT. This experiment is to confirm how much the low resources the model can handle. The results are shown in Table \ref{table:9}. When only 15.7 hours of labeled video-text data are used, it achieves 75.62\% WER, which shows the model cannot correctly learn from only 1/3 of the data. When 31.3 hours of data are utilized, the WER performance is 69.63\%. This performance is better than that of the other methods obtained using the full data in Table \ref{table:3}. The results indicate that the proposed method can perform well even with 2/3 of the data on mTEDx-IT by outperforming the previous methods trained on full data. By using 100\% of data (47h), the performance is improved to 68.04\% WER.

\textbf{Different amounts of audio-text data.} We also experiment with different amounts of audio-text data including the cases where the audio-text data is even smaller than the video-text data (47h). The results are shown in Table \ref{table:10}. The results indicate too small audio data (12h) leads to even worse performance than the randomly initialized decoder (\ie, 73.24\% WER). We found that when we use about 75\% amount (35h) of the video-text data, it starts to improve the performance. By using more audio-text data, we can improve the performance more. When using 147h and 294h of the audio-text paired data, we achieve 70.3\% and 68.0\% WERs, respectively, on mTEDx-IT dataset.

\textbf{Performances of pre-trained ASR models.} We provide the performances of the pre-trained ASR models on each audio-text paired dataset, before being applied to the lip reading tasks. We also provide the performances of the pre-trained LMDecoders on each audio-text paired dataset, before being applied to the lip reading tasks. Different from the ASR models, LMDecoders are trained from quantized audio units with LM while the ASR models are trained from continuous audio. Please note the objective of pre-training the LMDecoder is for applying it to lip reading, not for performing ASR (\ie, Audio-based Speech Recognition). The WER(\%) results are shown in Table \ref{table:11}. As the results indicate, the ASR performances of LMDecoder do not perform better than the \textit{ASR Pre-train}. However, the lip reading performances for low-resource languages of the proposed LMDecoder outperform the \textit{ASR Pre-train} as shown in Tables \ref{table:3}, \ref{table:4}, \ref{table:5}, and \ref{table:6}. From the results, we can confirm that the proposed pre-training strategies are more suitable for low-resource language lip reading than just pre-training a decoder through ASR.

%########### Table 9 #############
\input{Table/Table9.tex}
%#################################

%########### Table 10 #############
\input{Table/Table10.tex}
%#################################

%########### Table 11 #############
\input{Table/Table11.tex}
%#################################

\section{Conclusion}
This paper proposed a novel lip reading framework for low-resource languages. To address the challenge of insufficient video-text paired data of low-resource languages, we proposed to learn and combine general speech knowledge and language-specific knowledge. Specifically, the visual encoder is trained with masked predictions of speech units to learn general speech knowledge, and Language-specific Memory-augmented Decoder (LMDecoder) is proposed to learn language-specific knowledge from audio-text paired data. By combining the learned general speech knowledge and language-specific knowledge, we can efficiently develop lip reading models for low-resource languages. Through comprehensive experiments on a total of five languages (English, Italian, French, Spanish, and Portuguese), we verified the effectiveness of the proposed lip reading framework in low-resource languages. 

\section{Acknowledgment}
This work was partly supported by two funds: the National Research Foundation of Korea (NRF) grant funded by the Korea government (MSIT) (No. NRF-2022R1A2C2005529) and IITP grant funded by the Korea government(MSIT) (No.2020-0-00004, Development of Previsional Intelligence based on Long-Term Visual Memory Network)

{\small
\bibliographystyle{ieee_fullname}
\bibliography{egbib}
}

\appendix
\section{Training Details}
\subsection{LRS2}
Our proposed method outperforms the previous state-of-the-art methods and sets a new state-of-the-art performance in Table \ref{table:2}. In this section, we provide further details for pre-training the proposed LMDecoder. In addition, we also give details for finetuning the entire lip reading model with the LMDecoder on LRS2 dataset.

\subsubsection{Pre-training}
We pre-train the LMDecoder to learn English-specific knowledge from 656 hours of audio-text paired data of LRS2 and LRS3. The LMDecoder consists of LM, transformer encoders, and transformer decoders. We set the memory bank size as 1,000 and use 4 layers for transformer encoders with a 1,024 embedding dimension, a 4,096 feed-forward dimension, and 8 attention heads. The configuration of the transformer decoders is the same as transformer encoders except for having 9 layers. All components of the LMDecoder are trained in an end-to-end manner with 60,000 steps. We use warmup steps of 15,000. The other training options such as learning rate are shown in Table \ref{table:s1}. The tri-learning rate schedule in the table indicates (warmup, hold, decay) percentage for the total steps.

\subsubsection{Finetuning}
After the pre-training stage, we compose the lip reading pipeline by concatenating the pre-trained visual encoder and the LMDecoder. We employ a pre-trained AV-HuBERT Large model for the visual encoder. The entire lip reading model is finetuned for 30,000 steps. During finetuning, we freeze the visual encoder until 20,000 steps. Adam optimizer with a peak learning rate of 0.0005 and warmup steps of 10,000 is utilized for finetuning. Details are provided in the last column of Table \ref{table:s1}.

%########### Table s1 #############
\input{Table/Table_s1.tex}
%#################################

%########### Table s2 #############
\input{Table/Table_s2.tex}
%#################################

\subsection{mTEDx}
For the low-resource languages, our goal is to learn language-specific knowledge on each target language by using audio-text paired data to supplement insufficient video-text paired data. Therefore, we jointly utilize mTEDx and MLS datasets to pre-train LMDecoder on the target language data. Please note that the MLS dataset has more audio-text paired data than the mTEDx.

\subsubsection{Pre-training}
We pre-train the LMDecoder to learn language-specific knowledge from audio-text paired data of each target language (IT: 294h, FR: 1,163h, ES: 992h, and PT: 254h). The LMDecoder consists of LM, transformer encoders, and transformer decoders. We set the memory bank size as 1,000 and use 4 layers for transformer encoders with a 768 embedding dimension, a 3,072 feed-forward dimension, and 12 attention heads. The configuration of the transformer decoders is the same as transformer encoders except for having 6 layers. All components of the LMDecoder are trained in an end-to-end manner with 60,000 steps. We use warmup steps of 15,000. The other training options such as learning rate are shown in Table \ref{table:s2}.

\subsubsection{Finetuning}
After the pre-training stage, we compose the lip reading pipeline by concatenating the pre-trained visual encoder and the LMDecoder for each target language. We employ a pre-trained AV-HuBERT Base model for the visual encoder. The entire lip reading model is finetuned for 50,000 steps. In contrast to the experiment on LRS2, we do not freeze the visual encoder. Adam optimizer with a peak learning rate of 0.001 and warmup steps of 10,000 is utilized for finetuning. Details are provided in the last column of Table \ref{table:s2}.

\section{Utilizing Speech Knowledge of Large-scale Pre-trained English Lip Reading Model}
Recently, large-scale pre-trained lip reading models using ASR-labeled English data have been proposed \cite{ma2023auto}. They utilized a pre-trained ASR model to label unlabeled English datasets and obtained 3,448 hours of visual-text data. In this section, we explore whether we can utilize these large-scale pre-trained English lip reading models' speech knowledge for low-resource lip reading. However, as the visual encoder of their pre-trained model is not trained with the speech unit prediction task, it is not matched well with the LMDecoder that is trained using speech unit inputs. Therefore, we find the performance degradation when directly cascading the pre-trained visual encoder of \cite{ma2023auto} and the LMDecoder. To handle this, we add a residual connection between the output of the visual encoder and the input of the decoder so that the imperfect memory addressing in Language-specific Memory (LM) can be complemented through the residual connection. With this simple modification, we applied the proposed method to combine the speech knowledge learned from large-scale English data and the language-specific knowledge learned from language-specific audio-text data. The results on mTEDx are shown in Table \ref{table:s3}, \ref{table:s4}, \ref{table:s5}, and \ref{table:s6}. Compared to using 814 hours of English data, by employing the knowledge learned from 3,448 hours of English data, we can largely improve the lip reading performances for low-resource languages (\ie, IT, FR, ES, and PT). These results confirm that the speech knowledge learned from one language can be transferred to different languages. By combining speech knowledge with language-specific knowledge through the proposed method, we can further improve lip reading performances on low-resource language datasets. For example, we can improve about 3\% WER more from that of \cite{ma2023auto} on the mTEDx-ES dataset.
%########### Table s3 ############
\input{Table/Table_s3.tex}
%#################################

%########### Table s4 ############
\input{Table/Table_s4.tex}
%#################################

%########### Table s5 ############
\input{Table/Table_s5.tex}
%#################################

%########### Table s6 ############
\input{Table/Table_s6.tex}
%#################################

\end{document}

%% file: Figure/Figure_1.tex
\begin{figure*}[t]
	\centering
	\centerline{\includegraphics[width=18.5cm]{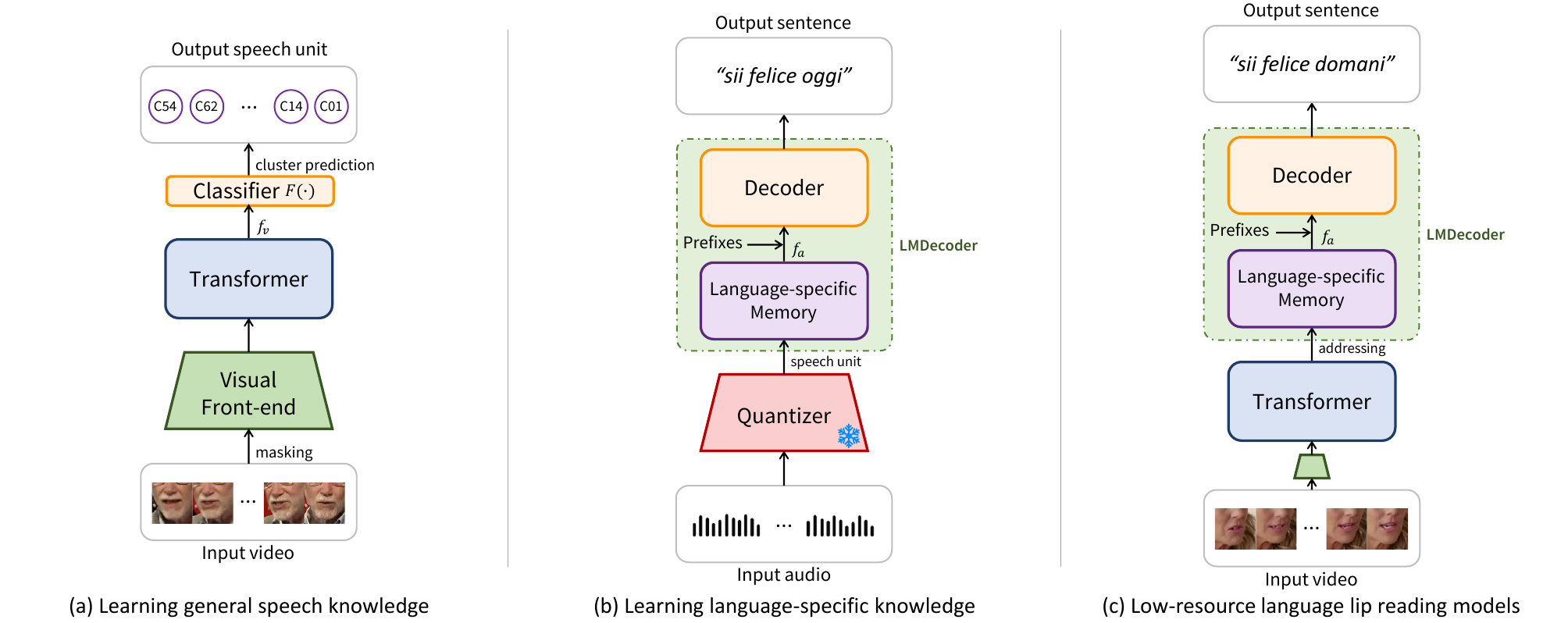}}
	\caption{Overview of the proposed method for low-resource language lip reading. (a) Learning general speech representation by using masked prediction of speech units in a high-resource language. (b) The proposed Language-specific Memory-augmented Decoder (LMDecoder) learns language-specific knowledge from audio-text paired data by quantizing the input into speech units. (c) Lip reading models for low-resource languages can be built by combining general speech knowledge and language-specific knowledge.
	}
	\label{fig:1}
\end{figure*}

%% file: Figure/Figure_2.tex
\begin{figure}[t]
	\centering
	\centerline{\includegraphics[width=7.5cm]{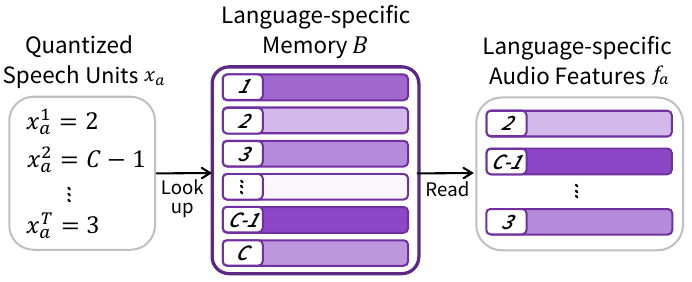}}
	\caption{Illustration of Language-specific Memory (LM) and the memory banks $B$ of LM. When a quantized speech unit is given, LM transforms it into a language-specific audio feature by reading the memory value. Therefore, the mapping of speech units to language-specific audio features can be constructed.
	}
	\label{fig:2}
\end{figure}

%% file: Table/Table1.tex
\begin{table}[t]
	\renewcommand{\arraystretch}{1.2}
	\renewcommand{\tabcolsep}{4mm}
\centering
\resizebox{0.999\linewidth}{!}{
\begin{tabular}{ccccc}
\Xhline{3\arrayrulewidth}
\textbf{Modality} & \textbf{Dataset}  & \textbf{Train} & \textbf{Validation} & \textbf{Test} \\ \hline
A-T & MLS-ES & 918 & 10 & 10 \\
A-T & MLS-FR & 1,077 & 10 & 10 \\
A-T & MLS-IT & 247 & 5 & 5 \\
A-T & MLS-PT & 161 & 4 & 4 \\ \hdashline
V-T & mTEDx-ES & 74 & 0.7 & 0.5 \\
V-T & mTEDx-FR & 86 & 0.4 & 0.3 \\
V-T & mTEDx-IT & 47 & 0.4 & 0.4 \\
V-T & mTEDx-PT & 93 & 0.7 & 0.7 \\
\Xhline{3\arrayrulewidth}
\end{tabular}}
\vspace{0.05cm}
\caption{Data length (Hours) of each dataset. A-T represents audio-text paired data and V-T represents video-text paired data.}
\label{table:1}
\end{table}

%% file: Table/Table2.tex
\begin{table}[t]
	\renewcommand{\arraystretch}{1.2}
	\renewcommand{\tabcolsep}{9.5mm}
\centering
\resizebox{0.85\linewidth}{!}{
\begin{tabular}{cc}
\Xhline{3\arrayrulewidth}
\textbf{Method}  & \textbf{WER (\%)} \\ \hline
Afouras \etal \cite{afouras2020asr}  & 58.5 \\
Zhang \etal \cite{zhang2019spatio} & 51.7 \\
TM-seq2seq \cite{afouras2018deep} & 48.3 \\
CroMM-VSR \cite{kim2021cromm} & 46.2 \\
MVM \cite{kim2022distinguishing} & 44.5 \\
CM-seq2seq \cite{ma2021end} & 37.9 \\
Prajwal \etal \cite{prajwal2022sub} & 28.9 \\
Auxiliary Task \cite{ma2022multiple} & 28.7 \\
AV-HuBERT \cite{shi2022learning} & 25.5 \\
VATLM \cite{zhu2023vatlm} & 24.3 \\\hline
\textbf{Proposed Method} & \textbf{23.8} \\
\Xhline{3\arrayrulewidth}
\end{tabular}}
\vspace{+0.1cm}
\caption{Comparisons with state-of-the-art methods on LRS2.}
\label{table:2}
\end{table}

%% file: Table/Table3.tex
\begin{table}[t]
	\renewcommand{\arraystretch}{1.4}
	\renewcommand{\tabcolsep}{1mm}
\centering
\resizebox{0.9999\linewidth}{!}{
\begin{tabular}{ccccc}
\Xhline{3\arrayrulewidth}
\textbf{Method} & \makecell{\textbf{Unlabeled}\\ \textbf{V-A Data}} & \makecell{\textbf{Labeled}\\ \textbf{A-T Data}} & \makecell{\textbf{Labeled}\\ \textbf{V-T Data}}  & \textbf{WER} \\ \hline
CM-seq2seq \cite{ma2021end} & - & - & 47h (+814h) & 78.31\% \\
AV-HuBERT \cite{shi2022learning} & 1759h  & - & 47h & 73.24\% \\
ASR Pre-train & 1759h & 294h & 47h & 71.28\% \\
Auxiliary Task \cite{ma2022multiple} & 1759h & 47h  & 47h & 71.99\% \\
\hline
\textbf{Proposed Method} & 1759h &  294h & 47h & \textbf{68.04\%} \\
\Xhline{3\arrayrulewidth}
\end{tabular}}
\vspace{0.05cm}
\caption{Lip reading performance comparisons on mTEDx-IT. (+$\alpha$) represents the amount of labeled English data.}
\label{table:3}
\end{table}

%% file: Table/Table4.tex
\begin{table}[t]
	\renewcommand{\arraystretch}{1.4}
	\renewcommand{\tabcolsep}{1mm}
\centering
\resizebox{0.9999\linewidth}{!}{
\begin{tabular}{ccccc}
\Xhline{3\arrayrulewidth}
\textbf{Method} & \makecell{\textbf{Unlabeled}\\ \textbf{V-A Data}} & \makecell{\textbf{Labeled}\\ \textbf{A-T Data}} & \makecell{\textbf{Labeled}\\ \textbf{V-T Data}}  & \textbf{WER} \\ \hline
CM-seq2seq \cite{ma2021end} & - & - & 86h (+814h) & 88.41\% \\
AV-HuBERT \cite{shi2022learning} & 1759h & - & 86h & 77.36\% \\
ASR Pre-train & 1759h & 1163h  & 86h & 75.67\% \\
Auxiliary Task \cite{ma2022multiple} & 1759h & 86h  & 86h & 76.79\% \\
\hline
\textbf{Proposed Method} & 1759h & 1163h  & 86h & \textbf{74.74\%} \\
\Xhline{3\arrayrulewidth}
\end{tabular}}
\vspace{0.05cm}
\caption{Lip reading performance comparisons on mTEDx-FR. (+$\alpha$) represents the amount of labeled English data.}
\label{table:4}
\end{table}

%% file: Table/Table5.tex
\begin{table}[t]
	\renewcommand{\arraystretch}{1.4}
	\renewcommand{\tabcolsep}{1mm}
\centering
\resizebox{0.9999\linewidth}{!}{
\begin{tabular}{ccccc}
\Xhline{3\arrayrulewidth}
\textbf{Method} & \makecell{\textbf{Unlabeled}\\ \textbf{V-A Data}} & \makecell{\textbf{Labeled}\\ \textbf{A-T Data}} & \makecell{\textbf{Labeled}\\ \textbf{V-T Data}}  & \textbf{WER} \\ \hline
CM-seq2seq \cite{ma2021end} & - & - & 74h (+814h) & 81.75\% \\
AV-HuBERT \cite{shi2022learning} & 1759h & - & 74h & 71.68\% \\
ASR Pre-train & 1759h &  992h & 74h & 70.80\% \\
Auxiliary Task \cite{ma2022multiple} & 1759h & 74h  & 74h & 70.91\% \\ 
\hline
\textbf{Proposed Method} & 1759h & 992h  & 74h & \textbf{70.16\%} \\
\Xhline{3\arrayrulewidth}
\end{tabular}}
\vspace{0.05cm}
\caption{Lip reading performance comparisons on mTEDx-ES. (+$\alpha$) represents the amount of labeled English data.}
\label{table:5}
\end{table}

%% file: Table/Table6.tex
\begin{table}[t]
	\renewcommand{\arraystretch}{1.4}
	\renewcommand{\tabcolsep}{1mm}
\centering
\resizebox{0.9999\linewidth}{!}{
\begin{tabular}{ccccc}
\Xhline{3\arrayrulewidth}
\textbf{Method} & \makecell{\textbf{Unlabeled}\\ \textbf{V-A Data}} & \makecell{\textbf{Labeled}\\ \textbf{A-T Data}} & \makecell{\textbf{Labeled}\\ \textbf{V-T Data}}  & \textbf{WER} \\ \hline
CM-seq2seq \cite{ma2021end} & - & - & 93h (+814h) & 79.17\% \\
AV-HuBERT \cite{shi2022learning} & 1759h & - & 93h & 71.87\% \\
ASR Pre-train & 1759h & 254h & 93h & 70.39\% \\
Auxiliary Task \cite{ma2022multiple} & 1759h & 93h &  93h & 70.39\% \\
\hline
\textbf{Proposed Method} & 1759h & 254h & 93h & \textbf{69.33\%} \\
\Xhline{3\arrayrulewidth}
\end{tabular}}
\vspace{0.05cm}
\caption{Lip reading performance comparisons on mTEDx-PT. (+$\alpha$) represents the amount of labeled English data.}
\label{table:6}
\end{table}

%% file: Table/Table7.tex
\begin{table}[t]
	\renewcommand{\arraystretch}{1.4}
	\renewcommand{\tabcolsep}{3.5mm}
\centering
\resizebox{0.9999\linewidth}{!}{
\begin{tabular}{cccc}
\Xhline{3\arrayrulewidth}
\makecell{\textbf{Train data}\\ \textbf{for LMDecoder}} & \makecell{\textbf{Labeled}\\ \textbf{A-T Data}} & \makecell{\textbf{Labeled}\\ \textbf{V-T Data}}  & \textbf{WER} \\ \hline
Baseline Decoder & 0h & 47h & 73.24\% \\ \hdashline
MLS-IT & 247h & 47h & 70.45\% \\
mTEDx-IT & 47h & 47h & 71.47\% \\
\textbf{MLS-IT+mTEDx-IT} & 294h & 47h & \textbf{68.04\%} \\
\Xhline{3\arrayrulewidth}
\end{tabular}}
\vspace{0.05cm}
\caption{Ablation study using different audio-text paired data.}
\label{table:7}
\end{table}

%% file: Table/Table8.tex
\begin{table}[t]
	\renewcommand{\arraystretch}{1.4}
	\renewcommand{\tabcolsep}{2.8mm}
\centering
\resizebox{0.9999\linewidth}{!}{
\begin{tabular}{ccccc}
\Xhline{3\arrayrulewidth}
\textbf{Method} & \makecell{\textbf{Unlabeled}\\ \textbf{V-A Data}} & \makecell{\textbf{Labeled}\\ \textbf{A-T Data}} & \makecell{\textbf{Labeled}\\ \textbf{V-T Data}}  & \textbf{WER} \\ \hline
Without LM & 1759h & 293h & 46h & 71.01\% \\
\textbf{With LM} & 1759h & 293h & 46h & \textbf{68.04\%} \\
\Xhline{3\arrayrulewidth}
\end{tabular}}
\vspace{0.05cm}
\caption{Ablation study with and without LM on mTEDx-IT.}
\label{table:8}
\end{table}

%% file: Table/Table9.tex
\begin{table}[t]
	\renewcommand{\arraystretch}{1.4}
	\renewcommand{\tabcolsep}{3.8mm}
\centering
\resizebox{0.9999\linewidth}{!}{
\begin{tabular}{cccc}
\Xhline{3\arrayrulewidth}
\textbf{V-T Data Amount} & \textbf{15.7h} & \textbf{31.3h} & \textbf{47h} \\ \hline
\textbf{WER} & 75.62\% & 69.63\% & 68.04\% \\
\Xhline{3\arrayrulewidth}
\end{tabular}
}
\vspace{0.05cm}
\caption{Ablation study using different amounts of video-text paired data on mTEDx-IT.}
\label{table:9}
\end{table}

%% file: Table/Table10.tex
\begin{table}[t]
	\renewcommand{\arraystretch}{1.4}
	\renewcommand{\tabcolsep}{1.7mm}
\centering
\resizebox{0.9999\linewidth}{!}{
\begin{tabular}{cccccc}
\Xhline{3\arrayrulewidth}
\textbf{A-T Data Amount} & \textbf{12h} & \textbf{35h} & \textbf{47h} & \textbf{147h} & \textbf{297h} \\ \hline
\textbf{WER} & 86.7\% & 72.2\% & 71.5\% & 70.3\% & 68.0\% \\
\Xhline{3\arrayrulewidth}
\end{tabular}
}
\vspace{0.05cm}
\caption{Ablation study using different amounts of audio-text paired data on mTEDx-IT.}
\label{table:10}
\end{table}

%% file: Table/Table11.tex
\begin{table}[t]
	\renewcommand{\arraystretch}{1.4}
	\renewcommand{\tabcolsep}{7mm}
\centering
\resizebox{0.9999\linewidth}{!}{
\begin{tabular}{ccc}
\Xhline{3\arrayrulewidth}
\textbf{Dataset} & \textbf{ASR Pre-train} & \textbf{LMDecoder}  \\ \hline
\textbf{mTEDx-IT} & 24.65\% & 29.21\%  \\
\textbf{mTEDx-FR} & 22.96\% & 27.48\% \\
\textbf{mTEDx-ES} & 25.01\% & 24.65\% \\
\textbf{mTEDx-PT} & 28.35\% & 36.01\%  \\
\Xhline{3\arrayrulewidth}
\end{tabular}}
\vspace{0.05cm}
\caption{Performances of pre-trained models (ASR) on mTEDx.}
\label{table:11}
\end{table}

%% file: Table/Table_s1.tex
\begin{table}[t]
	\renewcommand{\arraystretch}{1.1}
	\renewcommand{\tabcolsep}{1.5mm}
\centering
\resizebox{1.0\linewidth}{!}{
\begin{tabular}{ccc}
\Xhline{3\arrayrulewidth}
& Pre-training & Fine-tuning  \\ \hline
\# steps & 60,000 & 30,000\\
\# frozen steps & - & 20,000 \\
tri-stage LR schedule & (25\%, 0\%, 75\%) & (33\%, 0\%, 67\%) \\
peak learning rate & 1e-3 & 5e-4  \\
\# GPUs & 8 & 8  \\
Adam $(\beta_{1}, \beta_{2})$ & (0.9, 0.98) & (0.9, 0.98)  \\
\Xhline{3\arrayrulewidth}
\end{tabular}}
\vspace{0.1cm}
\caption{Training details on LRS2 (EN).}
\label{table:s1}
\end{table}

%% file: Table/Table_s2.tex
\begin{table}[t]
	\renewcommand{\arraystretch}{1.1}
	\renewcommand{\tabcolsep}{1.5mm}
\centering
\resizebox{1.0\linewidth}{!}{
\begin{tabular}{ccc}
\Xhline{3\arrayrulewidth}
& Pre-training & Fine-tuning  \\ \hline
\# steps & 60,000 & 50,000 \\
\# frozen steps & - & - \\
tri-stage LR schedule & (25\%, 0\%, 75\%) & (20\%, 0\%, 80\%) \\
peak learning rate & 1e-3 & 1e-3  \\
\# GPUs & 8 & 4  \\
Adam $(\beta_{1}, \beta_{2})$ & (0.9, 0.98) & (0.9, 0.98)  \\
\Xhline{3\arrayrulewidth}
\end{tabular}}
\vspace{0.1cm}
\caption{Training details on mTEDx (IT, FR, ES, and PT).}
\label{table:s2}
\end{table}

%% file: Table/Table_s3.tex
\begin{table}[t]
	\renewcommand{\arraystretch}{1.4}
	\renewcommand{\tabcolsep}{2.5mm}
\centering
\resizebox{0.9999\linewidth}{!}{
\begin{tabular}{cccc}
\Xhline{3\arrayrulewidth}
\textbf{Method} & \makecell{\textbf{Labeled}\\ \textbf{A-T Data}} & \makecell{\textbf{Labeled}\\ \textbf{V-T Data}}  & \textbf{WER} \\ \hline
CM-seq2seq \cite{ma2021end} &  - & 47h (+814h) & 78.31\%  \\
CM-seq2seq \cite{ma2023auto} &  - & 47h (+3448h) & 60.40\% \\
\hline
\textbf{Proposed Method} &  294h & 47h (+3448h) & \textbf{59.74\%}  \\
\Xhline{3\arrayrulewidth}
\end{tabular}}
\vspace{0.05cm}
\caption{Lip reading performance comparisons on mTEDx-IT. (+$\alpha$) represents the amount of labeled English data.}
\label{table:s3}

\end{table}

%% file: Table/Table_s4.tex
\begin{table}[t]
	\renewcommand{\arraystretch}{1.4}
	\renewcommand{\tabcolsep}{2.5mm}
\centering
\resizebox{0.9999\linewidth}{!}{
\begin{tabular}{cccc}
\Xhline{3\arrayrulewidth}
\textbf{Method} & \makecell{\textbf{Labeled}\\ \textbf{A-T Data}} & \makecell{\textbf{Labeled}\\ \textbf{V-T Data}}  & \textbf{WER} \\ \hline
CM-seq2seq \cite{ma2021end} &  - & 86h (+814h) & 88.41\%  \\
CM-seq2seq \cite{ma2023auto} &  - & 86h (+3448h) & 65.25\% \\
\hline
\textbf{Proposed Method} &  1163h & 86h (+3448h) & \textbf{64.92\%}  \\
\Xhline{3\arrayrulewidth}
\end{tabular}}
\vspace{0.05cm}
\caption{Lip reading performance comparisons on mTEDx-FR. (+$\alpha$) represents the amount of labeled English data.}
\label{table:s4}

\end{table}

%% file: Table/Table_s5.tex
\begin{table}[t]
	\renewcommand{\arraystretch}{1.4}
	\renewcommand{\tabcolsep}{2.5mm}
\centering
\resizebox{0.9999\linewidth}{!}{
\begin{tabular}{cccc}
\Xhline{3\arrayrulewidth}
\textbf{Method} & \makecell{\textbf{Labeled}\\ \textbf{A-T Data}} & \makecell{\textbf{Labeled}\\ \textbf{V-T Data}}  & \textbf{WER} \\ \hline
CM-seq2seq \cite{ma2021end} &  - & 74h (+814h) & 81.75\%  \\
CM-seq2seq \cite{ma2023auto} &  - & 74h (+3448h) & 59.90\% \\
\hline
\textbf{Proposed Method} &  992h & 74h (+3448h) & \textbf{56.96\%}  \\
\Xhline{3\arrayrulewidth}
\end{tabular}}
\vspace{0.05cm}
\caption{Lip reading performance comparisons on mTEDx-ES. (+$\alpha$) represents the amount of labeled English data.}
\label{table:s5}

\end{table}

%% file: Table/Table_s6.tex
\begin{table}[t]
	\renewcommand{\arraystretch}{1.4}
	\renewcommand{\tabcolsep}{2.5mm}
\centering
\resizebox{0.9999\linewidth}{!}{
\begin{tabular}{cccc}
\Xhline{3\arrayrulewidth}
\textbf{Method} & \makecell{\textbf{Labeled}\\ \textbf{A-T Data}} & \makecell{\textbf{Labeled}\\ \textbf{V-T Data}}  & \textbf{WER} \\ \hline
CM-seq2seq \cite{ma2021end} &  - & 93h (+814h) & 79.17\%  \\
CM-seq2seq \cite{ma2023auto} &  - & 93h (+3448h) & 59.45\% \\
\hline
\textbf{Proposed Method} &  254h & 93h (+3448h) & \textbf{58.57\%}  \\
\Xhline{3\arrayrulewidth}
\end{tabular}}
\vspace{0.05cm}
\caption{Lip reading performance comparisons on mTEDx-PT. (+$\alpha$) represents the amount of labeled English data.}
\label{table:s6}

\end{table}